\documentclass[lettersize,journal]{IEEEtran}
\pdfoutput=1
\usepackage{amsmath,amsfonts}
\usepackage{algorithm}
\usepackage{array}
\usepackage{textcomp}
\usepackage{stfloats}
\usepackage{url}
\usepackage{verbatim}
\usepackage{graphicx}
\usepackage{cite}
\usepackage{booktabs}
\usepackage{multirow}
\usepackage{makecell}
\usepackage{stackengine}
\usepackage{threeparttable}
\usepackage{xcolor} 
\usepackage{tikz}
\usepackage{soul}
\usepackage[normalem]{ulem}
\usepackage{diagbox}
\usepackage{subfig}
\usepackage{esvect}
\usepackage{algpseudocode}  
\usepackage[colorlinks,
bookmarksopen,
bookmarksnumbered,
citecolor=blue,
linkcolor=red,
urlcolor=red]{hyperref}
\usepackage{bookmark}
\usepackage{arydshln}
\usepackage{stfloats}
\hypersetup{pdfborder={0 0 0}}

\hypersetup{hidelinks,
	colorlinks=true,
	allcolors=black,
	pdfstartview=Fit,
	breaklinks=true}
	
\definecolor{lime}{HTML}{A6CE39}

\DeclareRobustCommand{\orcidicon}{	
\begin{tikzpicture}
\draw[lime, fill=lime] (0,0)
circle[radius=0.16]
node[white]{{\fontfamily{qag}\selectfont \tiny \.{I}D}};
\end{tikzpicture}
\hspace{-2mm}
}

\foreach \x in {A, ..., Z}{%
\expandafter\xdef\csname orcid\x\endcsname{\noexpand\href{https://orcid.org/\csname orcidauthor\x\endcsname}{\noexpand\orcidicon}}
}

\begin{document}

\title{MomentBA: Second-order Spatial Moments for Anisotropic Correspondence Uncertainty in Differentiable Bundle Adjustment}

\author{
	 Yuqing Wang\hspace{-1.5mm}\orcidA{},~\IEEEmembership{Graduate Student Member, IEEE}, Xiaoji Niu\hspace{-1.5mm}\orcidF{}, ~\IEEEmembership{Member,~IEEE}, Yan Wang*\hspace{-1.5mm}\orcidB{}, Hailiang Tang\hspace{-1.5mm}\orcidC{}, Jian Kuang\hspace{-1.5mm}\orcidH{},and Tisheng Zhang\hspace{-1.5mm}\orcidG{},~\IEEEmembership{Member,~IEEE}
	\thanks{Yuqing Wang is with the GNSS Research Center, Wuhan University,	Wuhan 430079, China, and also with the Electronic Information School, Wuhan University, Wuhan 430079, China (e-mail: m18071357432@163.com).}
	\thanks{Tisheng Zhang and Xiaoji Niu are with the GNSS Research Center, Wuhan University, Wuhan 430079, China, also with the Electronic Information School, Wuhan University, Wuhan 430079, China, also with Hubei Technology Innovation Center for Spatiotemporal Information and Positioning Navigation, Wuhan 430079, China, and also with Hubei Luojia Laboratory, Wuhan 430079, China (e-mail: zts@whu.edu.cn; xjniu@whu.edu.cn).}
	\thanks{Yan Wang, Hailiang Tang, Jian Kuang, are with the GNSS Research Center, Wuhan University, Wuhan 430079, China, also with the Hubei Technology Innovation Center for Spatiotemporal Information and Positioning Navigation (e-mail: wystephen@whu.edu.cn, thl@whu.edu.cn, kuang@whu.edu.cn).}
}

\markboth{Journal of \LaTeX\ Class Files,~Vol.~14, No.~8, August~2021}%
{Shell \MakeLowercase{\textit{et al.}}: A Sample Article Using IEEEtran.cls for IEEE Journals}

\IEEEpubid{}

\maketitle

\begin{figure*}[t]
	\centering
	{\includegraphics[width=1\linewidth]{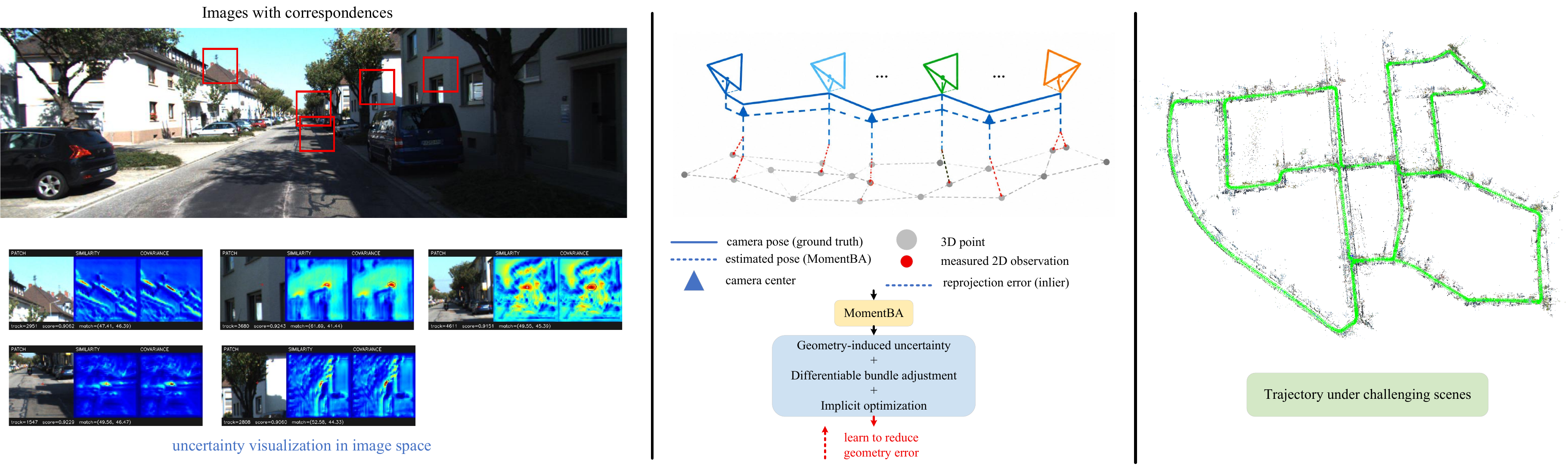}}
	
	\caption{We present \textbf{MomentBA}, a differentiable bundle adjustment framework that models anisotropic correspondence uncertainty from similarity responses via second-order spatial moments. The left shows correspondence images and uncertainty visualization, the middle illustrates geometry-induced uncertainty estimation and its integration into bundle adjustment, and the right demonstrates improved trajectory estimation under challenging scenes.}
	\label{fig:total}
\end{figure*}

\begin{abstract}
	Most existing visual odometry (VO) systems treat feature correspondences as deterministic measurements or assign uniform uncertainty, ignoring the inherent localization ambiguity of different observations. However, correspondence uncertainty is often anisotropic due to image structures such as edges, repetitive patterns, and motion blur, which can significantly affect geometric optimization. In this work, we propose MomentBA, a geometry-aware bundle adjustment framework that derives anisotropic correspondence uncertainty from second-order spatial moments of local similarity responses. Instead of introducing additional covariance prediction networks, the proposed method directly converts matching response distributions into interpretable covariance estimates and incorporates them into bundle adjustment as correspondence-specific information matrices for uncertainty-aware residual weighting. Furthermore, the proposed formulation is integrated into a differentiable optimization framework, establishing a direct connection between correspondence uncertainty and geometric estimation. Experiments on the EuRoC MAV and TartanAir v1 Hard datasets demonstrate that MomentBA improves monocular visual odometry accuracy compared with existing feature-based and learning-based approaches. The proposed anisotropic covariance model achieves lower rotational errors and more robust trajectory estimation than fixed and isotropic uncertainty models, validating the effectiveness of geometry-induced uncertainty modeling for challenging visual environments.
\end{abstract}


\section{Introduction}\label{I}

\IEEEPARstart{V}{isual} odometry (VO), visual--inertial odometry (VIO), and simultaneous localization and mapping (SLAM) estimate camera motion by minimizing geometric residuals over sparse feature correspondences~\cite{orb2,orb3,vins,dso}. Recent advances in learned feature detectors and matchers have significantly improved correspondence accuracy and robustness under challenging imaging conditions~\cite{SuperPoint,r2d2,aliked,lightglue}. Nevertheless, once feature correspondences are established, most existing VO/VIO systems either treat image observations as deterministic measurements or adopt simplified uncertainty models, ignoring the correspondence-specific localization ambiguity. Such an assumption neglects an important characteristic of visual measurements: different correspondences inherently exhibit different uncertainty distributions that directly affect downstream geometric optimization.

In real-world scenes, correspondence uncertainty is highly structured and anisotropic. Corner-like features usually provide well-constrained localization, whereas edge structures, repetitive textures, and motion blur often produce ambiguous observations that are accurately localized in one direction but highly uncertain in the orthogonal direction~\cite{stereo_uncertainty,uncertainty_hypervolume}. Such anisotropic uncertainty directly affects nonlinear estimation, since bundle adjustment relies on residual weighting according to measurement reliability. Ignoring directional uncertainty causes unreliable correspondences to contribute excessive constraints along ambiguous directions, resulting in degraded convergence and reduced localization accuracy.

Existing uncertainty-aware visual estimation methods attempt to address this problem by introducing probabilistic observation models or learning correspondence uncertainty from data~\cite{dnls,macvo}. Although these approaches demonstrate that uncertainty-aware optimization can improve pose estimation, the uncertainty is typically obtained through predefined models or additional prediction mechanisms, making it less directly connected to the intrinsic spatial structure of feature matching. In contrast, modern feature matching networks naturally produce dense local similarity responses whose spatial distributions encode correspondence ambiguity.Sharp responses indicate reliable localization, whereas elongated responses reveal directional uncertainty and multi-modal responses indicate matching ambiguity caused by competing correspondence hypotheses. However, this rich geometric information is usually discarded after selecting the maximum response as the final correspondence.

In this work, we argue that correspondence uncertainty should be regarded as a geometric property of the matching process rather than an external quantity requiring additional prediction. We introduce the concept of \emph{Geometry-induced Uncertainty (GIU)}, where correspondence covariance is directly derived from the local matching distribution. Specifically, the similarity response generated by a feature matcher is interpreted as a spatial probability distribution, and its second-order spatial moments provide an analytical characterization of anisotropic localization uncertainty. This formulation establishes a direct connection between image-level matching geometry and probabilistic measurement modeling, while preserving physical interpretability without introducing additional covariance estimation networks.

Based on this formulation, we propose \textbf{MomentBA}, a differentiable bundle adjustment framework that incorporates geometry-induced correspondence covariance into nonlinear geometric optimization, as shown in Fig.~\ref{fig:total}. Each correspondence is associated with an individual anisotropic information matrix, allowing reprojection residuals to be weighted according to both the magnitude and direction of localization uncertainty. Furthermore, the covariance formulation is coupled with differentiable bundle adjustment, enabling geometric consistency to provide supervision for the underlying matching representation. In this way, MomentBA establishes a unified framework that connects correspondence formation, uncertainty modeling, and geometric optimization, improving the robustness and accuracy of visual odometry under challenging environments.

The main contributions of this work are summarized as follows.

\begin{itemize}
	
\item \textbf{Geometry-induced correspondence uncertainty.}
We propose a novel uncertainty formulation that analytically derives anisotropic correspondence covariance from second-order spatial moments of local similarity responses, providing an explicit geometric interpretation of localization ambiguity.

\item \textbf{Moment-based anisotropic bundle adjustment.}
We formulate a geometry-aware bundle adjustment framework in which each correspondence is associated with an individual anisotropic information matrix, enabling uncertainty-aware residual weighting during nonlinear optimization.

\item \textbf{Comprehensive experimental validation.}
Extensive experiments validate the effectiveness of the proposed anisotropic uncertainty modeling for robust and accurate visual odometry.
	
\end{itemize}

The remainder of this paper is organized as follows. Section~\ref{II} reviews related work on uncertainty modeling and geometric optimization for visual odometry and SLAM. Section~\ref{III} introduces the feature extraction and matching representation used in our framework. Section~\ref{IV} presents the proposed geometry-induced anisotropic bundle adjustment framework, including covariance estimation and uncertainty-aware optimization. Section~\ref{V} provides experimental evaluations, including comparisons with state-of-the-art methods and comprehensive ablation studies. Finally, Section~\ref{VI} concludes the paper.

\section{Related Work}\label{II}

Reliable correspondence uncertainty is critical for visual odometry (VO), visual--inertial odometry (VIO), and SLAM, since the confidence assigned to feature observations directly affects geometric optimization. Existing studies have investigated uncertainty from different perspectives, including handcrafted measurement models, learned covariance prediction, and differentiable geometric optimization. However, most existing methods still do not explicitly derive correspondence uncertainty from the geometry of the matching process itself.

Early probabilistic VO methods estimate observation uncertainty from image gradients, tracking residuals, or predefined sensor noise models. Ross \emph{et al.}~\cite{stereo_uncertainty} studied uncertainty estimation for stereo visual odometry and demonstrated the importance of propagating feature-level uncertainty into pose estimation. Förstner and Gülch~\cite{dnls} analyzed feature localization uncertainty from local image structure, while later works investigated covariance modeling for KLT tracking and local feature matching~\cite{shi1994good,Lucas,Tomasi,mikolajczyk2005performance}. These methods provide explicit uncertainty models, but they usually rely on handcrafted assumptions and cannot fully characterize complex matching ambiguity caused by weak texture, repetitive structures, motion blur, or illumination changes.

Recent learning-based methods attempt to estimate correspondence uncertainty directly from data. Kendall and Gal~\cite{uncertainties} introduced uncertainty modeling into deep visual prediction, inspiring subsequent uncertainty-aware perception methods. Muhle \emph{et al.}~\cite{muhle2023learning} proposed DNLS, which learns per-feature covariance by differentiating through nonlinear relative pose estimation. Qiu \emph{et al.}~\cite{macvo} proposed MAC-VO, where metrics-aware covariance is used for both feature selection and residual weighting in stereo VO. These methods demonstrate that uncertainty-aware optimization improves visual localization robustness. Nevertheless, their uncertainty is treated as a latent quantity to be learned or regressed, rather than being directly induced from the spatial structure of the matching distribution.

Differentiable geometric optimization further enables visual learning to be supervised by downstream geometric objectives. BA-Net~\cite{ba-net} embeds differentiable bundle adjustment into depth and motion learning. DeepFactors~\cite{DeepFactors}, DROID-SLAM~\cite{DROID-SLAM}, and DPVO~\cite{dpvo} further demonstrate that differentiable pose estimation and bundle adjustment can improve dense or semi-dense visual odometry. Other optimization layers, such as OptNet~\cite{optnet}, Deep Declarative Networks~\cite{gould2021deep}, and implicit differentiation frameworks~\cite{blondel2022efficient}, provide general tools for differentiating through optimization problems. However, these methods mainly focus on propagating gradients through solvers, while the observation uncertainty used inside the solver is usually fixed, isotropic, scalar-weighted, or externally predicted.

Our work is motivated by a different observation: correspondence uncertainty is fundamentally a geometric property of the local matching distribution. Instead of predicting covariance using an auxiliary network or learning it indirectly from pose supervision, MomentBA derives anisotropic correspondence covariance directly from the second-order spatial moments of local similarity responses. The resulting geometry-induced uncertainty is then embedded into an implicitly differentiable bundle adjustment framework, establishing a unified connection between correspondence formation, uncertainty modeling, and geometric optimization.

\begin{figure*}
	\centering
	{\includegraphics[width=1\linewidth]{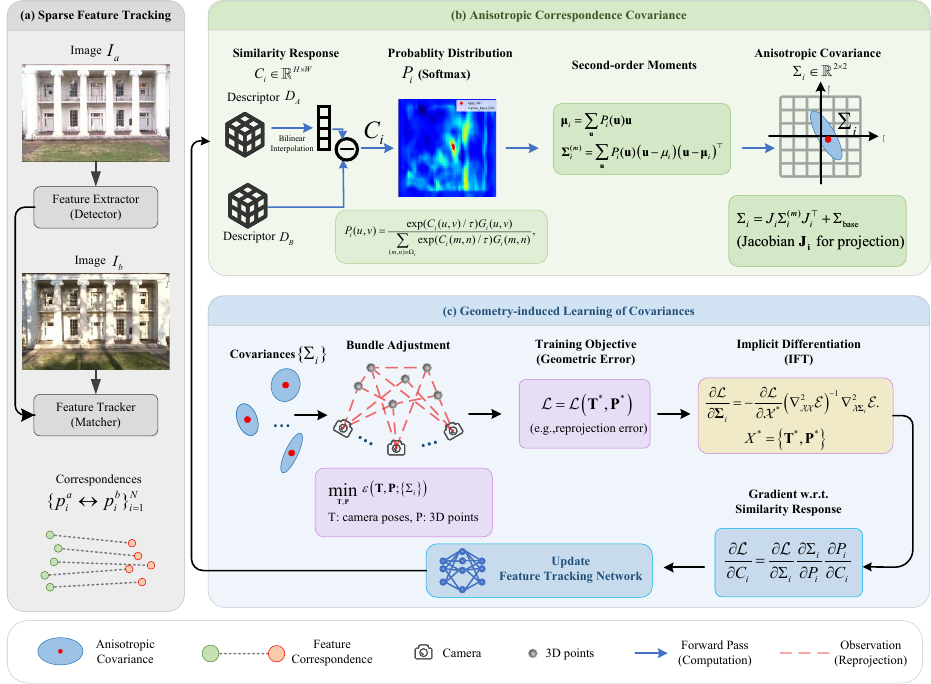}}
	
	\caption{Overview of MomentBA. Similarity responses from the tracking network are used to derive anisotropic covariances via spatial moments, which are incorporated into bundle adjustment for uncertainty-aware optimization and learning.}
	\label{fig:framework}
\end{figure*}

\section{FEATURE EXTRACTION AND MATCHING REPRESENTATION}
\label{III}

This section briefly introduces the sparse feature extraction and matching pipeline adopted in MomentBA. As illustrated in Fig.~\ref{fig:framework}, the front-end consists of a feature detector that identifies reliable sparse keypoints and a feature tracker that establishes inter-frame correspondences. The detector and tracker are adopted as standard front-end components \cite{viba} and are not the contribution of this work. Instead, our focus is the dense similarity response produced by the tracker, which serves as the basis for the geometry-induced covariance construction described in Sec.\ref{IV}.

\subsection{Sparse Feature Detection}

Given an input image
\(
\mathbf{I}\in\mathbb{R}^{H\times W\times3},
\)
the feature detector extracts a sparse set of reliable keypoints

\begin{equation}
	\mathcal{P}
	=
	\{\mathbf{p}_i\}_{i=1}^{N},
\end{equation}
where each keypoint corresponds to a stable and distinctive image location. Standard confidence estimation and non-maximum suppression are employed to remove redundant responses, resulting in a uniformly distributed set of sparse features. These keypoints serve as the reference observations for subsequent inter-frame matching.

\subsection{Similarity-based Feature Tracking}

For each reference keypoint
\(
\mathbf{p}_i\in\mathcal{P}
\)
in image
\(
\mathbf{I}_a,
\)
the feature tracker searches for its correspondence within a local search region
\(
\Omega_i
\)
of the target image
\(
\mathbf{I}_b.
\)

The reference descriptor is compared with all candidate descriptors inside the search region to construct a dense similarity response map

\begin{equation}
	\mathbf{C}_i
	\in
	\mathbb{R}^{H_s\times W_s},
\end{equation}
where each response value measures the similarity between the reference descriptor and the corresponding candidate location. The correspondence is obtained from the maximum (or sub-pixel refined maximum) of the similarity response.

Unlike conventional feature matching methods that only preserve the final correspondence location, MomentBA retains the complete similarity response map throughout the matching process. The spatial distribution of the response naturally characterizes the ambiguity of local correspondence estimation and therefore provides substantially richer geometric information than a single matching coordinate. Consequently, instead of treating the similarity response merely as an intermediate quantity for correspondence extraction, the proposed framework directly exploits its spatial statistics to construct anisotropic correspondence covariance, which is subsequently incorporated into geometry-induced bundle adjustment in Section~IV.

\section{Geometry-induced Anisotropic Bundle Adjustment}\label{IV}
In this section, we present the proposed geometry-induced anisotropic bundle adjustment framework. As shown in Fig \ref{fig:framework}(b) We first derive an interpretable anisotropic covariance from the spatial statistics of the local similarity response, providing an analytical characterization of correspondence uncertainty without relying on an auxiliary covariance prediction network. The derived covariance is subsequently incorporated into bundle adjustment as a correspondence-specific information matrix, where geometric optimization further refines the covariance through implicit differentiation. Consequently, as illustrated in Fig\ref{fig:framework}(c), the learned uncertainty is directly coupled with multi-view geometric consistency, enabling end-to-end optimization of the feature representation.

\subsection{Anisotropic Correspondence Covariance}

The similarity response produced by the tracking network contains richer information than the final correspondence location. Conventional feature matching methods retain only the response maximum as the estimated correspondence while discarding the remaining response distribution. However, the spatial shape of the similarity response naturally reflects the localization uncertainty of the correspondence. A compact response indicates a well-localized feature with high confidence, whereas elongated or flat responses usually arise from repetitive textures, motion blur, or edge structures, implying anisotropic localization ambiguity. Therefore, instead of learning covariance through an additional prediction network, we derive correspondence uncertainty directly from the spatial statistics of the similarity response.

Given the similarity response map $C_i(u,v)$ associated with the $i$-th correspondence over the local search region $\Omega_i$, let $\boldsymbol{\mu}_i$ denote the estimated correspondence location. To suppress distant secondary responses that may arise from repetitive structures, we introduce a Gaussian spatial window centered at $\boldsymbol{\mu}_i$,

\begin{equation}
	G_i(u,v)=
	\exp\left(
	-\frac{\|\mathbf{x}-\boldsymbol{\mu}_i\|_2^2}
	{2\sigma_w^2}
	\right),
	\label{eq:spatial_window}
\end{equation}
where $\mathbf{x}=[u,v]^T$ and $\sigma_w$ controls the spatial extent of the local response. The localized response is then normalized into a discrete probability distribution,

\begin{equation}
	P_i(u,v)=
	\frac{\exp(C_i(u,v)/\tau)G_i(u,v)}
	{\sum\limits_{(m,n)\in\Omega_i}
		\exp(C_i(m,n)/\tau)G_i(m,n)},
	\label{eq:probability}
\end{equation}
where $\tau$ controls the sharpness of the probability distribution and
$\sum_{(u,v)\in\Omega_i}P_i(u,v)=1$.

The localization uncertainty is characterized by the second-order spatial moment around the estimated correspondence,

\begin{equation}
	\mathbf{\Sigma}_i
	=
	\sum_{(u,v)\in\Omega_i}
	P_i(u,v)
	(\mathbf{x}-\boldsymbol{\mu}_i)
	(\mathbf{x}-\boldsymbol{\mu}_i)^{T}.
	\label{eq:covariance}
\end{equation}

Expanding Eq.~(\ref{eq:covariance}) gives

\begin{equation}
	\mathbf{\Sigma}_i=
	\begin{bmatrix}
		\sigma_{u}^{2} & \sigma_{uv}\\
		\sigma_{uv} & \sigma_{v}^{2}
	\end{bmatrix},
	\label{eq:covariance_matrix}
\end{equation}
whose diagonal terms represent the localization variance along the horizontal and vertical directions, while the off-diagonal term describes the correlation between the two axes.

Finally, eigenvalue decomposition is applied to the covariance matrix,

\begin{equation}
	\mathbf{\Sigma}_i
	=
	\mathbf{Q}_i
	\mathbf{\Lambda}_i
	\mathbf{Q}_i^{T},
	\label{eq:eigen}
\end{equation}
where $\mathbf{\Lambda}_i=\mathrm{diag}(\lambda_1,\lambda_2)$ contains the principal variances and $\mathbf{Q}_i=[\mathbf{q}_1,\mathbf{q}_2]$ defines the corresponding principal directions. Large eigenvalues indicate high localization uncertainty, while the associated eigenvectors reveal the dominant ambiguity direction. For numerical stability, the eigenvalues are bounded within $[\lambda_{\min},\lambda_{\max}]$ before reconstructing the covariance matrix.

The resulting covariance provides an analytical and interpretable characterization of anisotropic correspondence uncertainty and is subsequently incorporated into bundle adjustment as a correspondence-specific information matrix for anisotropic residual weighting.

\subsection{Geometry-induced Learning of Covariances}

The covariance derived from the similarity response characterizes the local uncertainty of each correspondence from image appearance alone. However, local matching statistics do not explicitly account for multi-view geometric consistency. To further improve the uncertainty estimation, we incorporate the proposed covariance into bundle adjustment and optimize it under geometric supervision. Consequently, the covariance is no longer determined solely by local similarity distributions, but is continuously refined according to the global reprojection objective.

Let $\mathcal{X}=\{\mathbf{T},\mathbf{P}\}$ denote the optimization variables consisting of camera poses $\mathbf{T}$ and 3D landmarks $\mathbf{P}$. Given the anisotropic covariance $\mathbf{\Sigma}_i$ associated with the $i$-th correspondence, its corresponding information matrix is

\begin{equation}
	\mathbf{W}_i=\mathbf{\Sigma}_i^{-1}.
	\label{eq:information}
\end{equation}

The reprojection residual of the $i$-th observation is defined as

\begin{equation}
	\mathbf{r}_i
	=
	\mathbf{u}_i
	-
	\pi(\mathbf{T}_{c(i)},\mathbf{P}_{j(i)}),
	\label{eq:reprojection}
\end{equation}
where $\pi(\cdot)$ denotes the camera projection model, $\mathbf{u}_i$ is the observed feature location, while $c(i)$ and $j(i)$ indicate the corresponding camera and landmark indices, respectively.

Instead of minimizing the conventional isotropic reprojection error, we formulate the geometry-aware bundle adjustment objective using the Mahalanobis distance,

\begin{equation}
	\mathcal{E}(\mathcal{X},\mathbf{\Sigma})
	=
	\sum_i
	\rho
	\!\left(
	\mathbf{r}_i^{T}
	\mathbf{W}_i
	\mathbf{r}_i
	\right),
	\label{eq:ba_energy}
\end{equation}
where $\rho(\cdot)$ denotes the Huber robust kernel. Compared with isotropic least squares, the proposed objective assigns an individual anisotropic confidence to each correspondence, allowing ambiguous observations to contribute less along poorly constrained directions while preserving strong constraints in reliable directions.

The optimal geometric state is obtained by solving

\begin{equation}
	\mathcal{X}^{*}
	=
	\arg\min_{\mathcal{X}}
	\mathcal{E}(\mathcal{X},\mathbf{\Sigma}).
	\label{eq:ba_opt}
\end{equation}

Since the covariance directly determines the information matrix in Eq.~(\ref{eq:ba_energy}), the optimization result is inherently dependent on the estimated uncertainty. Consequently, minimizing the geometric objective provides supervision for covariance estimation, enabling correspondence uncertainty to be learned from multi-view geometric consistency rather than solely from local appearance.

\subsection{Implicit Backpropagation for Covariance Learning}

Since the optimal solution in Eq.~(\ref{eq:ba_opt}) is implicitly defined by the covariance-dependent bundle adjustment objective, gradients cannot be obtained by directly back-propagating through the iterative optimization process. Instead, following the implicit function theorem (IFT), we differentiate the first-order optimality condition of the optimization problem to compute the gradient of the converged solution with respect to the correspondence covariance.

Let the optimization variables be denoted by
\[
\mathcal{X}=\{\mathbf{T},\mathbf{P}\},
\]
where $\mathbf{T}$ and $\mathbf{P}$ represent the camera poses and 3D landmarks, respectively. The optimal solution satisfies

\begin{equation}
	\mathcal{X}^{*}
	=
	\arg\min_{\mathcal{X}}
	\mathcal{E}(\mathcal{X},\mathbf{\Sigma}),
	\label{eq:ift_opt}
\end{equation}
whose first-order optimality condition is

\begin{equation}
	\nabla_{\mathcal{X}}
	\mathcal{E}
	(\mathcal{X}^{*},\mathbf{\Sigma})
	=
	\mathbf{0}.
	\label{eq:optimality}
\end{equation}

Differentiating Eq.~(\ref{eq:optimality}) with respect to the correspondence covariance yields

\begin{equation}
	\nabla_{\mathcal{X}\mathcal{X}}^{2}\mathcal{E}
	\,
	\frac{\partial\mathcal{X}^{*}}
	{\partial\mathbf{\Sigma}}
	+
	\nabla_{\mathcal{X}\mathbf{\Sigma}}^{2}\mathcal{E}
	=
	\mathbf{0},
	\label{eq:ift1}
\end{equation}
where
$\nabla_{\mathcal{X}\mathcal{X}}^{2}\mathcal{E}$
denotes the Hessian matrix of the bundle adjustment objective with respect to the optimization variables, and
$\nabla_{\mathcal{X}\mathbf{\Sigma}}^{2}\mathcal{E}$
is the mixed second-order derivative.

According to the implicit function theorem,

\begin{equation}
	\frac{\partial\mathcal{X}^{*}}
	{\partial\mathbf{\Sigma}}
	=
	-
	\left(
	\nabla_{\mathcal{X}\mathcal{X}}^{2}\mathcal{E}
	\right)^{-1}
	\nabla_{\mathcal{X}\mathbf{\Sigma}}^{2}\mathcal{E}.
	\label{eq:ift2}
\end{equation}

Given the training objective
$\mathcal{L}(\mathcal{X}^{*})$,
the gradient with respect to the correspondence covariance is obtained by

\begin{equation}
	\frac{\partial\mathcal{L}}
	{\partial\mathbf{\Sigma}}
	=
	\frac{\partial\mathcal{L}}
	{\partial\mathcal{X}^{*}}
	\frac{\partial\mathcal{X}^{*}}
	{\partial\mathbf{\Sigma}}
	=
	-
	\frac{\partial\mathcal{L}}
	{\partial\mathcal{X}^{*}}
	\left(
	\nabla_{\mathcal{X}\mathcal{X}}^{2}\mathcal{E}
	\right)^{-1}
	\nabla_{\mathcal{X}\mathbf{\Sigma}}^{2}\mathcal{E}.
	\label{eq:ift4}
\end{equation}
\begin{figure*}[t]
	\centering
	{\includegraphics[width=1\linewidth]{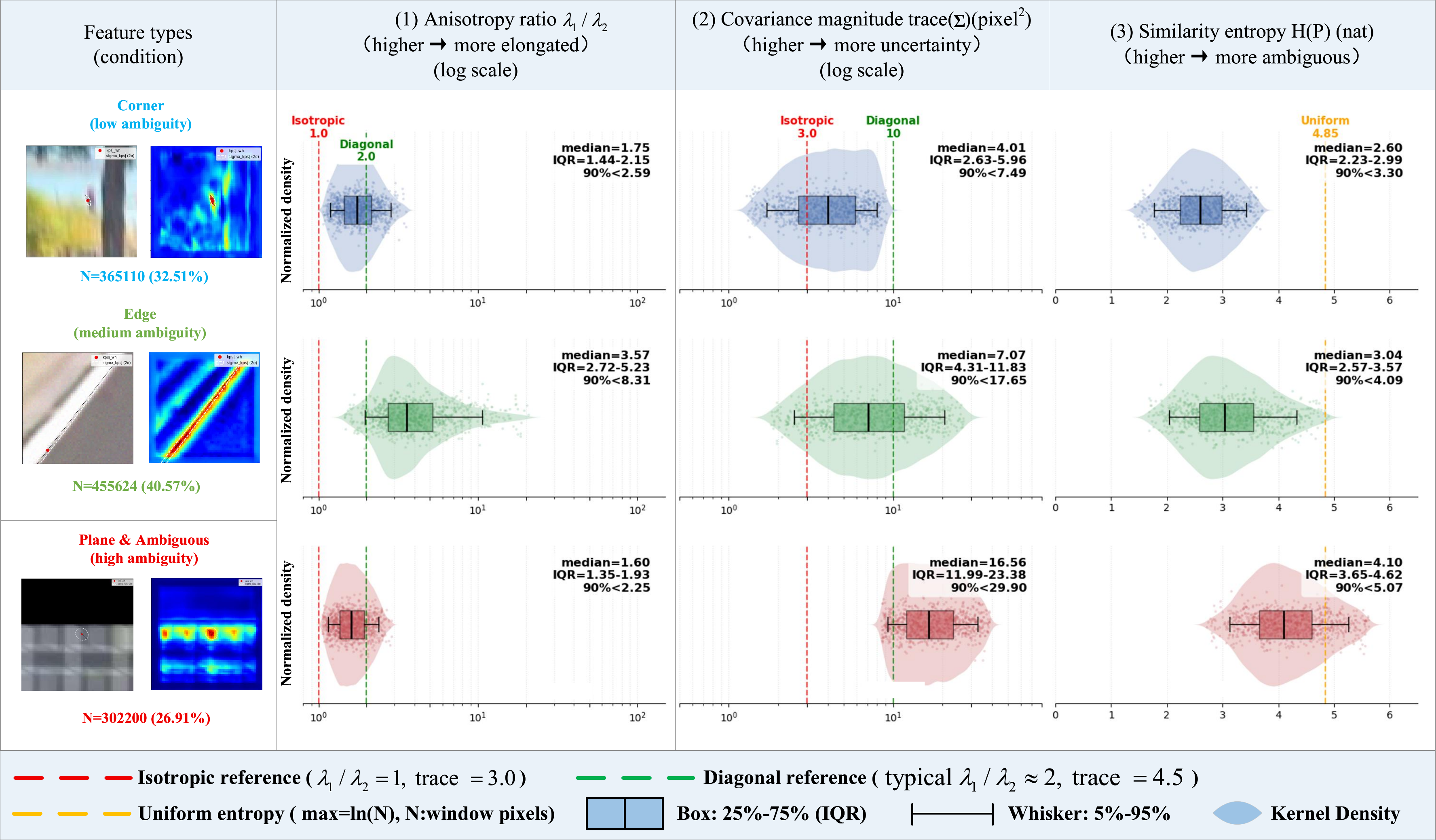}}
	
\caption{Statistics of MomentBA covariance across feature types. Distributions of anisotropy ratio $\lambda_1/\lambda_2$, covariance trace $\mathrm{tr}(\mathbf{\Sigma})$, and similarity entropy $H(P)$ show increasing geometric ambiguity from corners to edges and repetitive regions.}

	\label{fig:statics}
\end{figure*}
Since the proposed anisotropic covariance is analytically computed from the similarity response through the second-order spatial moment formulation, the covariance gradient in Eq.~(\ref{eq:ift4}) is further back-propagated through the chain rule as

\begin{equation}
	\frac{\partial\mathcal{L}}{\partial\theta}
	=
	\frac{\partial\mathcal{L}}{\partial\mathbf{\Sigma}}
	\frac{\partial\mathbf{\Sigma}}{\partial P}
	\frac{\partial P}{\partial\mathbf{C}}
	\frac{\partial\mathbf{C}}{\partial\theta},
	\label{eq:backward}
\end{equation}
where $\mathbf{C}$ denotes the similarity response map, $P$ is the corresponding probability distribution obtained by Eq.~(\ref{eq:probability}), and $\theta$ represents the parameters of the feature tracking network.

Consequently, the feature tracking network is encouraged to generate similarity responses whose induced anisotropic covariance minimizes the global bundle adjustment objective, establishing a direct connection between local matching ambiguity and multi-view geometric consistency.

\section{EXPERIMENTS}\label{V}
\subsection{Validation of Geometry-Aware Covariance Estimation}

We first evaluate whether the proposed MomentBA covariance reflects the intrinsic ambiguity of visual measurements. Three representative feature conditions are considered: distinctive corners, edge structures, and repetitive ambiguous regions. For each feature type, we analyze the distributions of anisotropy ratio $\lambda_1/\lambda_2$, covariance magnitude $\mathrm{tr}(\mathbf{\Sigma})$, and similarity entropy $H(P)$. For all experiments, the Gaussian window standard deviation is fixed to $\sigma_w=15$ pixels.

As shown in Fig.~\ref{fig:statics}, MomentBA covariance exhibits clear geometry-dependent characteristics. For corner features, the estimated uncertainty remains compact and nearly isotropic, with an anisotropy ratio close to the isotropic reference ($\lambda_1/\lambda_2=1$) and a small covariance trace (median $\mathrm{tr}(\mathbf{\Sigma})=4.01$). In contrast, edge features introduce strong directional ambiguity, leading to increased anisotropy (median $\lambda_1/\lambda_2=3.57$) and larger uncertainty magnitude (median $\mathrm{tr}(\mathbf{\Sigma})=7.07$). For repetitive structures, the covariance magnitude increases substantially (median $\mathrm{tr}(\mathbf{\Sigma})=16.56$), accompanied by higher similarity entropy ($H(P)=4.10$), indicating severe ambiguity in local correspondence localization. Although the anisotropy is lower than edge features ($\lambda_1/\lambda_2=1.60$), the increased entropy and covariance magnitude correctly represent the lack of confidence caused by repetitive patterns.

These results demonstrate that MomentBA covariance is not only adaptive to feature geometry but also consistent with the underlying uncertainty characteristics of visual correspondence, providing a more reliable measurement model for subsequent bundle adjustment.

\begin{table*}[t]
	\centering
	\caption{Relative pose error comparison on EuRoC dataset.
		$t_{rel}$ denotes translational error and $r_{rel}$ denotes rotational error.}
	\label{tab:euroc}
	\resizebox{\textwidth}{!}{
		\begin{tabular}{ccccccccccccccc}
			\toprule
			
			\multirow{2}{*}{Method}
			& \multicolumn{2}{c}{MH03}
			& \multicolumn{2}{c}{MH05}
			& \multicolumn{2}{c}{V102}
			& \multicolumn{2}{c}{V103}
			& \multicolumn{2}{c}{V202}
			& \multicolumn{2}{c}{V203}
			& \multicolumn{2}{c}{Avg.}
			\\
			
			\cmidrule(lr){2-3}
			\cmidrule(lr){4-5}
			\cmidrule(lr){6-7}
			\cmidrule(lr){8-9}
			\cmidrule(lr){10-11}
			\cmidrule(lr){12-13}
			\cmidrule(lr){14-15}
			
			&
			$t_{rel}$ & $r_{rel}$
			&
			$t_{rel}$ & $r_{rel}$
			&
			$t_{rel}$ & $r_{rel}$
			&
			$t_{rel}$ & $r_{rel}$
			&
			$t_{rel}$ & $r_{rel}$
			&
			$t_{rel}$ & $r_{rel}$
			&
			$t_{rel}$ & $r_{rel}$
			\\
			
			\midrule
			
			
			\multicolumn{15}{l}{\textbf{VO}}
			\\
			
			\hspace{5mm}TartanVO
			&0.0302&0.2791
			&0.0193&0.0604
			&0.0251&0.1244
			&0.0263&0.1552
			&0.0171&0.1251
			&0.0303&0.2986
			&0.0247&0.1738
			\\
			
			\hspace{5mm}MAC-VO
			&\textbf{0.0023}&0.0238
			&\underline{0.0025}&0.0216
			&\textbf{0.0029}&0.0434
			&\textbf{0.0032}&0.0580
			&\textbf{0.0018}&0.0406
			&\textbf{0.0049}&0.1284
			&\textbf{0.0029}&0.0526
			\\

			\midrule
			
			
			\multicolumn{15}{l}{\textbf{Feature}}
			\\
			
			\hspace{5mm}ORB
			&0.0064&0.0200
			&0.0085&0.0147
			&0.0067&\underline{0.0406}
			&0.0186&0.0481
			&0.0061&0.0411
			&0.0123&0.0628
			&0.0098&0.0379
			\\

			\hspace{5mm}SuperPoint
			&0.0053&0.0199
			&0.0088&0.0147
			&0.0066&0.0410
			&0.0081&0.0439
			&0.0056&\underline{0.0325}
			&0.0279&0.0746
			&0.0104&0.0378
			\\
			
			\hspace{5mm}SuperPoint-DAC
			&0.0057&\underline{0.0184}
			&0.0250&0.0161
			&0.0051&0.0409
			&0.0072&\textbf{0.0406}
			&0.0050&0.0325
			&0.0172&0.0728
			&0.0109&0.0369
			\\

			\hspace{5mm}ALIKED
			&0.0046&0.0200
			&0.0050&\underline{0.0144}
			&0.0080&0.0408
			&0.0074&0.0439
			&0.0081&0.0441
			&0.0320&\underline{0.0519}
			&0.0109&\underline{0.0359}
			\\

			
			\multicolumn{15}{l}{\textbf{Ours}}
			\\

			\hspace{5mm}constant
			&0.0044&0.0207
			&0.0052&0.0155
			&0.0068&0.0453
			&0.0086&0.0472
			&0.0057&0.0501
			&0.0108&0.0785
			&0.0069&0.0429
			\\

			\hspace{5mm}anisotropic
			&\underline{0.0035}&\textbf{0.0179}
			&\textbf{0.0022}&\textbf{0.0133}
			&\underline{0.0037}&\textbf{0.0361}
			&\underline{0.0054}&\underline{0.0415}
			&\underline{0.0029}&\textbf{0.0310}
			&\underline{0.0093}&\textbf{0.0501}
			&\underline{0.0045}&\textbf{0.0317}
			\\
			
			\bottomrule
			
		\end{tabular}
	}
\end{table*}

\begin{table*}[t]
	\centering
	\caption{Relative pose error comparison on TartanAir v1 Hard dataset.
		$t_{rel}$ denotes translational error and $r_{rel}$ denotes rotational error.}
	\label{tab:result}
	\resizebox{\textwidth}{!}{
		\begin{tabular}{ccccccccccccccc}
			\toprule
			
			\multirow{2}{*}{Method}
			& \multicolumn{2}{c}{H01}
			& \multicolumn{2}{c}{H03}
			& \multicolumn{2}{c}{H05}
			& \multicolumn{2}{c}{H07}
			& \multicolumn{2}{c}{H09}
			& \multicolumn{2}{c}{H011}
			& \multicolumn{2}{c}{Avg.}
			\\
			
			\cmidrule(lr){2-3}
			\cmidrule(lr){4-5}
			\cmidrule(lr){6-7}
			\cmidrule(lr){8-9}
			\cmidrule(lr){10-11}
			\cmidrule(lr){12-13}
			\cmidrule(lr){14-15}
			
			&
			$t_{rel}$ & $r_{rel}$
			&
			$t_{rel}$ & $r_{rel}$
			&
			$t_{rel}$ & $r_{rel}$
			&
			$t_{rel}$ & $r_{rel}$
			&
			$t_{rel}$ & $r_{rel}$
			&
			$t_{rel}$ & $r_{rel}$
			&
			$t_{rel}$ & $r_{rel}$
			\\
			
			\midrule
			
			\multicolumn{15}{l}{\textbf{VO}}
			\\
			
			\hspace{5mm}TartanVO
			&0.2918&0.0877
			&0.3275&0.1391
			&0.3644&0.1899
			&0.2605&0.0338
			&0.6182&0.0731
			&0.2204&0.0287
			&0.3471&0.0921
			\\

			\hspace{5mm}MAC-VO
			&0.3440&0.1052
			&0.2507&0.0778
			&\textbf{0.1485}&0.0352
			&\textbf{0.0985}&0.0418
			&0.3848&0.0628
			&\underline{0.1292}&0.1414
			&\underline{0.2260}&0.0774
			\\

			\midrule
			
			\multicolumn{15}{l}{\textbf{Feature}}
			\\

			\hspace{5mm}ORB
			&\textbf{0.1163}&0.0667
			&0.2212&\underline{0.0241}
			&0.2464&0.0298
			&0.2613&\underline{0.0325}
			&0.3454&0.0416
			&0.1826&\underline{0.0438}
			&0.2289&0.0398
			\\

			\hspace{5mm}SuperPoint
			&0.2570&0.0402
			&0.1950&0.0252
			&0.2314&0.0312
			&0.3012&0.0473
			&\underline{0.3238}&0.0414
			&0.1755&0.0513
			&0.2473&0.0394
			\\
			
			\hspace{5mm}SuperPoint-DAC
			&0.2510&\underline{0.0401}
			&\textbf{0.1947}&0.0252
			&0.2504&0.0315
			&0.3458&0.0402
			&0.2671&0.0435
			&0.1560&0.0503
			&0.2442&\underline{0.0385}
			\\
			
			\hspace{5mm}ALIKED
			&0.1655&0.0463
			&0.2215&0.0286
			&0.2537&0.0322
			&0.3081&0.0344
			&0.4431&\underline{0.0385}
			&0.1733&0.0545
			&0.2609&0.0391
			\\

			\midrule
			
			\multicolumn{15}{l}{\textbf{Ours}}
			\\

			\hspace{5mm}constant
			&0.2504&0.0406
			&0.2582&0.0378
			&0.2466&\underline{0.0297}
			&0.3227&0.0397
			&0.3313&0.0424
			&0.1769&0.0530
			&0.2643&0.0405
			\\

			\hspace{5mm}anisotropic
			&\underline{0.1603}&\textbf{0.0396}
			&\underline{0.1993}&\textbf{0.0215}
			&\underline{0.2084}&\textbf{0.0264}
			&\underline{0.1627}&\textbf{0.0237}
			&\textbf{0.2344}&\textbf{0.0367}
			&\textbf{0.1286}&\textbf{0.0371}
			&\textbf{0.1823}&\textbf{0.0308}
			\\
			
			\bottomrule
			
		\end{tabular}
	}
\end{table*}

\begin{figure}[t]
	\centering
	{\includegraphics[width=1\linewidth]{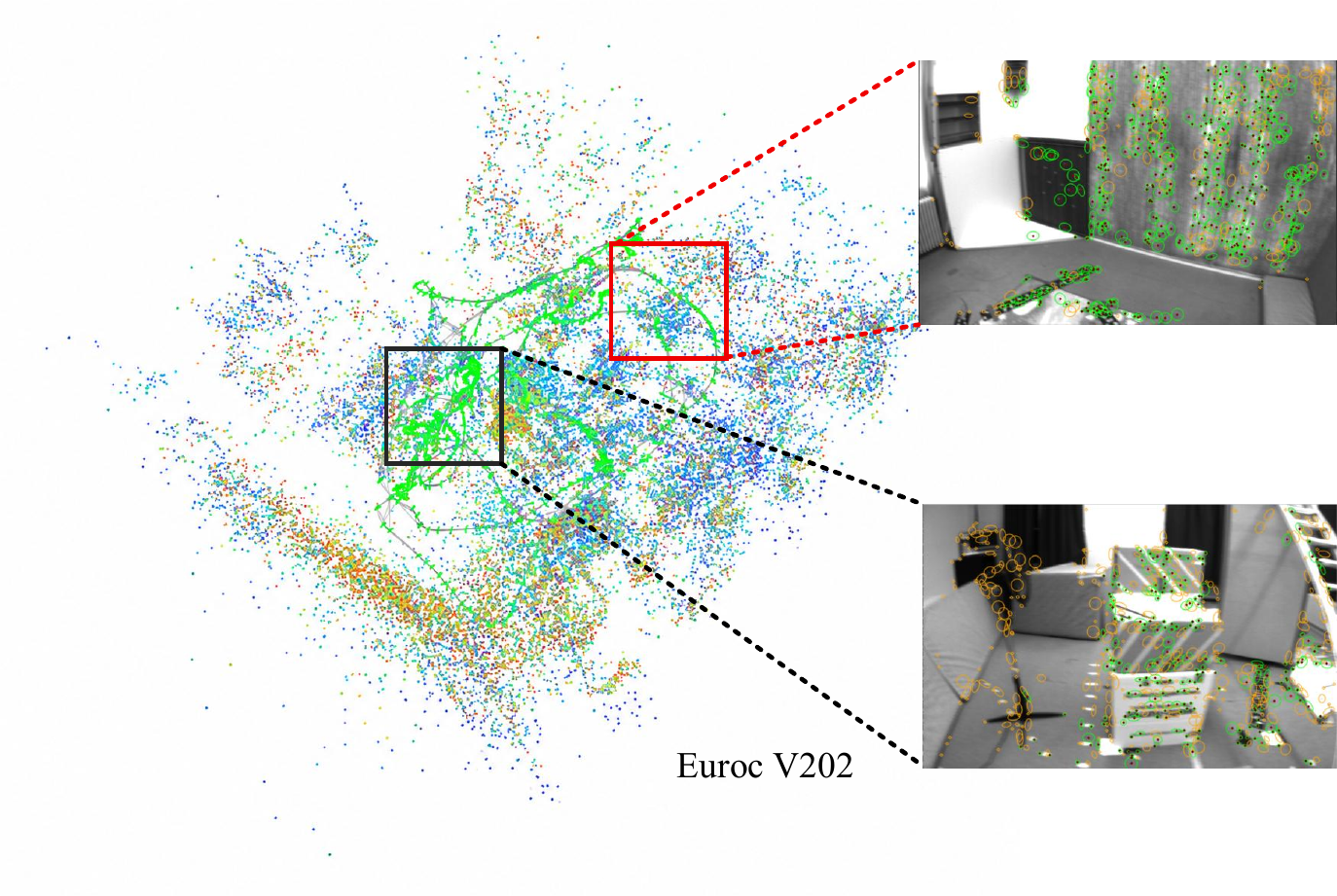}}
	
	\caption{Visualization of the estimated anisotropic correspondence covariance on the EuRoC MAV dataset.}
	\label{fig:traj(a)}
\end{figure}

\begin{figure}[t]
	\centering
	{\includegraphics[width=1\linewidth]{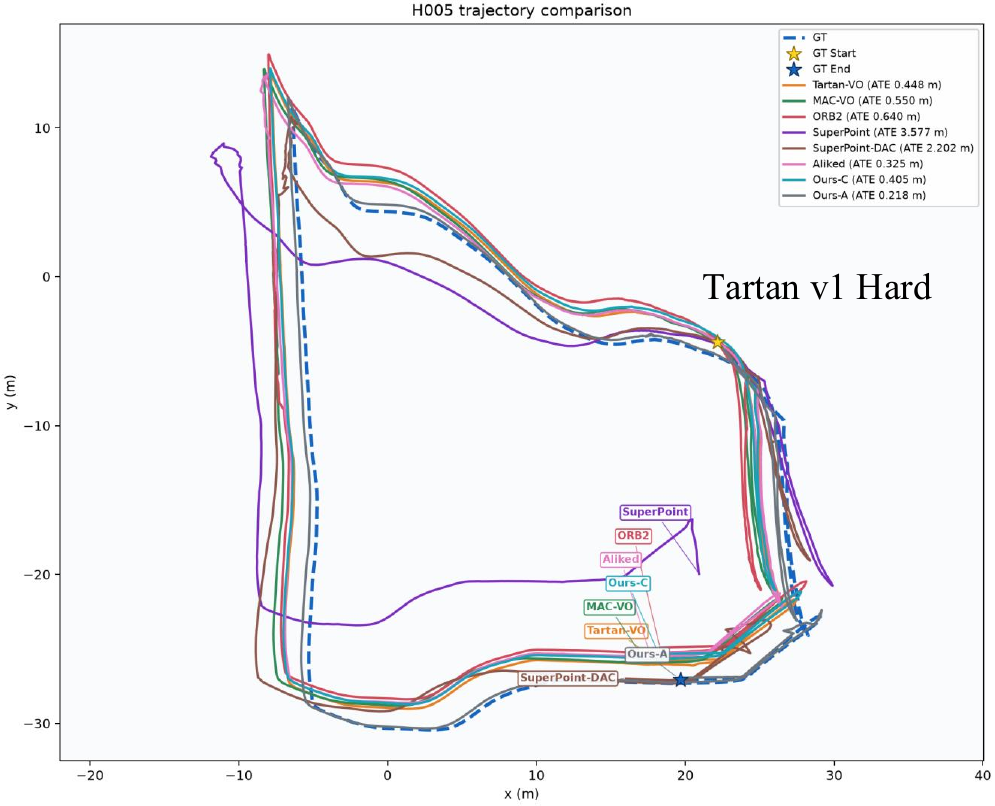}}
	
	\caption{Trajectory comparison of different methods on the TartanAir v1 Hard dataset.}
	\label{fig:traj(b)}
\end{figure}

\subsection{Trajectory Evaluation}

We evaluate the proposed monocular VO framework on the EuRoC MAV and TartanAir v1 Hard datasets using the open-source pyslam framework. The original ORB front-end is replaced with different feature representations, while loop closure is disabled. We compare feature-based methods (ORB, SuperPoint, and ALIKED) with learning-based VO approaches (TartanVO and MAC-VO). To further evaluate the effectiveness of uncertainty modeling, we additionally include SuperPoint-DAC, where the existing DAC\cite{dac} uncertainty estimation method is applied to SuperPoint correspondences. Relative pose error (RPE), including translational error $t_{rel}$ and rotational error $r_{rel}$, is used for evaluation. The results are summarized in Tab.~\ref{tab:euroc} and Tab.~\ref{tab:result}. 

On EuRoC, the proposed anisotropic covariance model achieves the best average performance, with $t_{rel}=0.0045$ and $r_{rel}=0.0317$. Compared with SuperPoint-DAC ($t_{rel}=0.0109$, $r_{rel}=0.0369$), our method reduces both translational and rotational errors, indicating that the proposed geometry-aware uncertainty modeling provides more effective constraints for bundle adjustment. It also reduces the average rotational error by 39.7\% compared with MAC-VO ($0.0526$ rad). On TartanAir v1 Hard, our method achieves average errors of $t_{rel}=0.1823$ and $r_{rel}=0.0308$, outperforming SuperPoint-DAC ($0.2442$ and $0.0385$, respectively) as well as the constant covariance setting ($0.2643$ and $0.0405$). The covariance distribution visualization on EuRoC and the trajectory comparison on TartanAir are shown in Fig.~\ref{fig:traj(a)} and Fig.~\ref{fig:traj(b)}, respectively. These results demonstrate that the proposed anisotropic covariance modeling consistently improves pose estimation and provides more effective uncertainty modeling for bundle adjustment across both real-world and synthetic challenging sequences.

\begin{table}[h]
	\centering
	\caption{Performance comparison of different ablation setups on the KITTI Dataset.}
	\label{tab:ablation}
	
	\scriptsize
	\setlength{\tabcolsep}{8pt}
	
	\resizebox{\linewidth}{!}{
		\begin{tabular}{c|ccccc}
			\toprule
			\multicolumn{6}{c}{Relative Translation Error ($t_{\mathrm{rel}}$, m/frame)}\\
			\midrule
			Trajectory & K01 & K03 & K05 & K07 & K09 \\
			\midrule
			constant
			&\underline{0.1089}&\underline{0.0321}&\underline{0.1820}&
			0.1252&\underline{0.1224}\\
			isotropic
			&0.2499&0.0344 & 0.1916 &\underline{0.0779}&0.1424\\
			anisotropic
			&\textbf{0.0802}&\textbf{0.0290}&\textbf{0.1394}&\textbf{0.0650}&\textbf{0.0996}\\
			\bottomrule
		\end{tabular}
	}
	
	\vspace{1mm}
	
	\resizebox{\linewidth}{!}{
		\begin{tabular}{c|ccccc}
			\toprule
			\multicolumn{6}{c}{Relative Rotation Error ($r_{\mathrm{rel}}$, $^\circ$/frame)}\\
			\midrule
			Trajectory & K01 & K03 & K05 & K07 & K09\\
			\midrule
			constant
			&\underline{0.4068}&0.0688&\underline{0.0516}&
			\underline{0.3037}&\underline{0.0802}\\
			isotropic
			&0.4297&\underline{0.0573}&\underline{0.0516}&0.2464&\underline{0.0802}\\
			anisotropic
			&\textbf{0.3839}&\textbf{0.0384}&\textbf{0.0352}&\textbf{0.1948}&\textbf{0.0433}\\
			\bottomrule
		\end{tabular}
	}
	
\end{table}

\subsection{Ablation Study}

To investigate different covariance modeling strategies, we conduct an ablation study with three variants: Constant, Isotropic, and Anisotropic covariance. Tab.~\ref{tab:result} compares the proposed anisotropic covariance with fixed and diagonal covariance models, while this section further analyzes different uncertainty representations.

The Constant variant assigns a fixed one-pixel covariance to all correspondences. The Isotropic variant estimates uncertainty magnitude but ignores directional information, whereas the Anisotropic variant models geometry-induced covariance from second-order spatial moments to preserve both uncertainty magnitude and directionality.

As shown in Tab.~\ref{tab:ablation}, the Anisotropic covariance achieves the lowest translation and rotation errors across all KITTI trajectories, demonstrating the effectiveness of directional uncertainty modeling. The Isotropic model provides adaptive uncertainty estimation but remains limited without directional information. By incorporating anisotropic uncertainty, the proposed covariance model provides more accurate constraints for bundle adjustment and improves trajectory estimation.

\section{Conclusion}\label{VI}

In this paper, we present MomentBA, a geometry-aware bundle adjustment framework that models anisotropic correspondence uncertainty from second-order spatial moments of local similarity responses. By converting the spatial distribution of matching responses into interpretable covariance estimates, MomentBA enables correspondence-specific uncertainty modeling and uncertainty-aware residual weighting during bundle adjustment.

The proposed formulation establishes a direct connection between feature correspondence ambiguity and geometric optimization without requiring additional covariance prediction networks. Extensive evaluations on the EuRoC MAV and TartanAir v1 Hard datasets demonstrate that incorporating geometry-induced anisotropic uncertainty improves monocular visual odometry accuracy and robustness. The proposed covariance model achieves consistent improvements over fixed and isotropic uncertainty models, validating the effectiveness of directional uncertainty modeling for challenging visual environments.

Overall, MomentBA provides a principled framework for integrating correspondence uncertainty into differentiable geometric optimization. Future work will explore extending the proposed uncertainty formulation to large-scale mapping and more complex multi-sensor visual localization systems.


\bibliographystyle{unsrt}
\bibliography{ref}

@misc{viba,
	title={ViBA: Implicit Bundle Adjustment with Geometric and Temporal Consistency for Robust Visual Matching}, 
	author={Xiaoji Niu and Yuqing Wang and Yan Wang and Hailiang Tang and Tisheng Zhang},
	year={2026},
	eprint={2604.03377},
	archivePrefix={arXiv},
	primaryClass={cs.CV},
	url={https://arxiv.org/abs/2604.03377}, 
}

@INPROCEEDINGS{dac,
	author={Tirado-Garín, Javier and Warburg, Frederik and Civera, Javier},
	booktitle={2024 International Conference on 3D Vision (3DV)}, 
	title={DAC: Detector-Agnostic Spatial Covariances for Deep Local Features}, 
	year={2024},
	volume={},
	number={},
	pages={728-738},
	doi={10.1109/3DV62453.2024.00034}}

@inproceedings{blondel2022efficient,
	author = {Blondel, Mathieu and Berthet, Quentin and Cuturi, Marco and Frostig, Roy and Hoyer, Stephan and Llinares-Lopez, Felipe and Pedregosa, Fabian and Vert, Jean-Philippe},
	booktitle = {Advances in Neural Information Processing Systems},
	editor = {S. Koyejo and S. Mohamed and A. Agarwal and D. Belgrave and K. Cho and A. Oh},
	pages = {5230--5242},
	publisher = {Curran Associates, Inc.},
	title = {Efficient and Modular Implicit Differentiation},
	url = {https://proceedings.neurips.cc/paper_files/paper/2022/file/228b9279ecf9bbafe582406850c57115-Paper-Conference.pdf},
	volume = {35},
	year = {2022}
}

@ARTICLE{gould2021deep,
	author={Gould, Stephen and Hartley, Richard and Campbell, Dylan},
	journal={IEEE Transactions on Pattern Analysis and Machine Intelligence}, 
	title={Deep Declarative Networks}, 
	year={2022},
	volume={44},
	number={8},
	pages={3988-4004},
	doi={10.1109/TPAMI.2021.3059462}}

@InProceedings{optnet,
	title = 	 {{O}pt{N}et: Differentiable Optimization as a Layer in Neural Networks},
	author =       {Brandon Amos and J. Zico Kolter},
	booktitle = 	 {Proceedings of the 34th International Conference on Machine Learning},
	pages = 	 {136--145},
	year = 	 {2017},
	editor = 	 {Precup, Doina and Teh, Yee Whye},
	volume = 	 {70},
	series = 	 {Proceedings of Machine Learning Research},
	month = 	 {06--11 Aug},
	publisher =    {PMLR},
	url = 	 {https://proceedings.mlr.press/v70/amos17a.html},
}

@INPROCEEDINGS{muhle2023learning,
	author={Muhle, Dominik and Koestler, Lukas and Jatavallabhula, Krishna Murthy and Cremers, Daniel},
	booktitle={2023 IEEE/CVF Conference on Computer Vision and Pattern Recognition (CVPR)}, 
	title={Learning Correspondence Uncertainty via Differentiable Nonlinear Least Squares}, 
	year={2023},
	volume={},
	number={},
	pages={13102-13112},
	doi={10.1109/CVPR52729.2023.01259}}

@misc{uncertainties,
	title={What Uncertainties Do We Need in Bayesian Deep Learning for Computer Vision?}, 
	author={Alex Kendall and Yarin Gal},
	year={2017},
	eprint={1703.04977},
	archivePrefix={arXiv},
	primaryClass={cs.CV},
	url={https://arxiv.org/abs/1703.04977}, 
}

@ARTICLE{mikolajczyk2005performance,
	author={Mikolajczyk, K. and Schmid, C.},
	journal={IEEE Transactions on Pattern Analysis and Machine Intelligence}, 
	title={A performance evaluation of local descriptors}, 
	year={2005},
	volume={27},
	number={10},
	pages={1615-1630},
	doi={10.1109/TPAMI.2005.188}}

@article{Tomasi,
	title={Detection and Tracking of Point Features},
	author={ Tomasi, C. },
	journal={Technical Report},
	volume={91},
	number={21},
	pages={9795-9802},
	year={1991},
}

@inproceedings{Lucas,
	title={An Iterative Image Registration Technique with an Application to Stereo Vision},
	author={Bruce D. Lucas and Takeo Kanade},
	booktitle={International Joint Conference on Artificial Intelligence},
	year={1981},
	url={https://api.semanticscholar.org/CorpusID:2121536}
}

@INPROCEEDINGS{shi1994good,
	author={Jianbo Shi and Tomasi},
	booktitle={1994 Proceedings of IEEE Conference on Computer Vision and Pattern Recognition}, 
	title={Good features to track}, 
	year={1994},
	volume={},
	number={},
	pages={593-600},
	doi={10.1109/CVPR.1994.323794}}

@INPROCEEDINGS{macvo,
	author={Qiu, Yuheng and Chen, Yutian and Zhang, Zihao and Wang, Wenshan and Scherer, Sebastian},
	booktitle={2025 IEEE International Conference on Robotics and Automation (ICRA)}, 
	title={MAC-VO: Metrics-Aware Covariance for Learning-Based Stereo Visual Odometry mac-vo.github.io}, 
	year={2025},
	volume={},
	number={},
	pages={3803-3814},
	doi={10.1109/ICRA55743.2025.11128482}}

@INPROCEEDINGS{dnls,
	author={Muhle, Dominik and Koestler, Lukas and Jatavallabhula, Krishna Murthy and Cremers, Daniel},
	booktitle={2023 IEEE/CVF Conference on Computer Vision and Pattern Recognition (CVPR)}, 
	title={Learning Correspondence Uncertainty via Differentiable Nonlinear Least Squares}, 
	year={2023},
	volume={},
	number={},
	pages={13102-13112},
	doi={10.1109/CVPR52729.2023.01259}}

@inproceedings{uncertainty_hypervolume,
	title={Uncertainty Hypervolume in Point Feature-Based Visual Odometry},
	author={InJun Mun and Sukhan Lee},
	booktitle={International Conference on Informatics in Control, Automation and Robotics},
	year={2024},
	url={https://api.semanticscholar.org/CorpusID:274225766}
}

@inproceedings{stereo_uncertainty,
	author    = {Ross, Derek and De Petrillo, Matteo and Strader, Jared and Gross, Jason N.},
	title     = {Uncertainty Estimation for Stereo Visual Odometry},
	booktitle = {Proceedings of the 34th International Technical Meeting of the Satellite Division of the Institute of Navigation (ION GNSS+ 2021)},
	year      = {2021},
	address   = {St. Louis, Missouri},
	month     = sep,
	pages     = {3263--3284},
	doi       = {10.33012/2021.18063}
}

@ARTICLE{vins,
	author={Qin, Tong and Li, Peiliang and Shen, Shaojie},
	journal={IEEE Transactions on Robotics}, 
	title={VINS-Mono: A Robust and Versatile Monocular Visual-Inertial State Estimator}, 
	year={2018},
	volume={34},
	number={4},
	pages={1004-1020},
	doi={10.1109/TRO.2018.2853729}}

@ARTICLE{orb2,
	author={Mur-Artal, Raúl and Tardós, Juan D.},
	journal={IEEE Transactions on Robotics}, 
	title={ORB-SLAM2: An Open-Source SLAM System for Monocular, Stereo, and RGB-D Cameras}, 
	year={2017},
	volume={33},
	number={5},
	pages={1255-1262},
	doi={10.1109/TRO.2017.2705103}}

@ARTICLE{orb3,
	author={Campos, Carlos and Elvira, Richard and Rodríguez, Juan J. Gómez and M. Montiel, José M. and D. Tardós, Juan},
	journal={IEEE Transactions on Robotics}, 
	title={ORB-SLAM3: An Accurate Open-Source Library for Visual, Visual–Inertial, and Multimap SLAM}, 
	year={2021},
	volume={37},
	number={6},
	pages={1874-1890},
	doi={10.1109/TRO.2021.3075644}}

@misc{dso,
	title={Direct Sparse Odometry}, 
	author={Jakob Engel and Vladlen Koltun and Daniel Cremers},
	year={2016},
	eprint={1607.02565},
	archivePrefix={arXiv},
	primaryClass={cs.CV},
	url={https://arxiv.org/abs/1607.02565}, 
}

@ARTICLE{aliked,
	author={Zhao, Xiaoming and Wu, Xingming and Chen, Weihai and Chen, Peter C. Y. and Xu, Qingsong and Li, Zhengguo},
	journal={IEEE Transactions on Instrumentation and Measurement}, 
	title={ALIKED: A Lighter Keypoint and Descriptor Extraction Network via Deformable Transformation}, 
	year={2023},
	volume={72},
	number={},
	pages={1-16},
	doi={10.1109/TIM.2023.3271000}}

@article{SuperPoint,
	title={SuperPoint: Self-Supervised Interest Point Detection and Description},
	author={Daniel DeTone and Tomasz Malisiewicz and Andrew Rabinovich},
	journal={2018 IEEE/CVF Conference on Computer Vision and Pattern Recognition Workshops (CVPRW)},
	year={2017},
	pages={337-33712},
	url={https://api.semanticscholar.org/CorpusID:4918026}
}

@misc{ba-net,
	title={BA-Net: Dense Bundle Adjustment Network}, 
	author={Chengzhou Tang and Ping Tan},
	year={2019},
	eprint={1806.04807},
	archivePrefix={arXiv},
	primaryClass={cs.CV},
	url={https://arxiv.org/abs/1806.04807}, 
}

@inbook{r2d2,
	author = {Revaud, Jerome and Weinzaepfel, Philippe and Souza, C\'{e}sar De and Humenberger, Martin},
	title = {R2D2: repeatable and reliable detector and descriptor},
	year = {2019},
	publisher = {Curran Associates Inc.},
	address = {Red Hook, NY, USA},
	booktitle = {Proceedings of the 33rd International Conference on Neural Information Processing Systems},
	articleno = {1113},
	numpages = {11}
}

@article{DeepFactors,
	title={DeepFactors: Real-Time Probabilistic Dense Monocular SLAM},
	volume={5},
	ISSN={2377-3774},
	url={http://dx.doi.org/10.1109/LRA.2020.2965415},
	DOI={10.1109/lra.2020.2965415},
	number={2},
	journal={IEEE Robotics and Automation Letters},
	publisher={Institute of Electrical and Electronics Engineers (IEEE)},
	author={Czarnowski, Jan and Laidlow, Tristan and Clark, Ronald and Davison, Andrew J.},
	year={2020},
	month=apr, pages={721–728} }

@misc{dpvo,
	title={Deep Patch Visual Odometry}, 
	author={Zachary Teed and Lahav Lipson and Jia Deng},
	year={2023},
	eprint={2208.04726},
	archivePrefix={arXiv},
	primaryClass={cs.CV},
	url={https://arxiv.org/abs/2208.04726}, 
}

@misc{DROID-SLAM,
	title={DROID-SLAM: Deep Visual SLAM for Monocular, Stereo, and RGB-D Cameras}, 
	author={Zachary Teed and Jia Deng},
	year={2022},
	eprint={2108.10869},
	archivePrefix={arXiv},
	primaryClass={cs.CV},
	url={https://arxiv.org/abs/2108.10869}, 
}

@INPROCEEDINGS{lightglue,
	author={Lindenberger, Philipp and Sarlin, Paul-Edouard and Pollefeys, Marc},
	booktitle={ICCV 2023}, 
	title={LightGlue: Local Feature Matching at Light Speed}, 
	year={2023},
	volume={},
	number={},
	pages={17581-17592},
}

\end{document}